# Who Sent It or How It Moved? Sender Shortcuts in Cryptocurrency Anti-Money-Laundering Benchmarks, Audited on Elliptic2 and Elliptic++

Mohammad Nasir Uddin
DBA-BIDA Program, Westcliff University, Irvine, CA, USA
m.uddin.258@westcliff.edu · ORCID 0009-0009-0990-4616

**Abstract**

Two public benchmarks dominate graph-based anti-money-laundering (AML) research on Bitcoin: the transaction-level Elliptic dataset, extended as Elliptic++, and the subgraph-level Elliptic2. In both, a label is assigned according to the entity that sends the funds. We ask whether models trained on these benchmarks learn how laundering moves or merely who sent the money, and we propose a diagnostic protocol to tell the two apart: a no-learning sender lookup, a novelty baseline, a sender-disjoint split with a leakage check, and a decomposition of performance into sender, receiver and intrinsic signal. On Elliptic2 the task largely reduces to classifying the sender. Under the standard random split, a lookup of how often a subgraph's senders fed suspicious subgraphs in training reaches a precision-recall area under the curve (PR-AUC) of 0.954, against 0.975 for gradient boosting on 456 features; our reimplementation of the best published method, RevClassify, reproduces its reported 0.97 and falls to 0.126 without sender information. When senders are held out, performance drops to 0.746 and the subgraph's own structure, transactions and receivers fall from 0.619 to 0.186. On Elliptic++, the same protocol finds the shortcut negligible. The corresponding sender-history baseline reaches only 0.340 against 0.986 for the full model (3.5 against 10.1 times the test prevalence), adding it to the full model under the random split raises PR-AUC by less than 0.004 in every seed, and the transaction's own features keep 0.883 when sending addresses are held out. We trace the difference to the unit of sender identity: Elliptic2 senders are entity clusters that recur across many labeled subgraphs, whereas Elliptic++ senders are individual addresses, and fewer than half of test transactions have a sending address seen in training. A shared sender-based label rule therefore produces a severe shortcut in one benchmark and a negligible one in the other, which is why the diagnostic must be run per benchmark. Code for both audits will be released.

## 1 Introduction

Investigators and compliance teams increasingly rely on machine learning to flag suspicious flows of cryptocurrency, and public benchmarks shape which methods are believed to work. Two families of benchmarks dominate. The Elliptic dataset (Weber et al., 2019) labels about 46,000 Bitcoin transactions as licit or illicit within a graph of 203,769 transactions, and Elliptic++ (Elmougy & Liu, 2023) extends it with the wallet addresses that send and receive each transaction. Elliptic2 (Bellei et al., 2024) moves to a larger scale and a different task: within a background graph of about 49 million entity clusters, it labels 121,810 small subgraphs, reflecting the view that laundering unfolds as a chain of transfers from a criminal source toward a service that converts the proceeds.

Results on Elliptic2 have improved quickly. The dataset paper reported a test PR-AUC of 0.208 for its best model, GLASS (Bellei et al., 2024). A later study, RevTrack (Song et al., 2024), reported 0.97 by representing each subgraph only through its initial senders and final receivers. Such a jump invites a basic question for any forensic use of these models: what, exactly, has been learned?

Our starting point is how the labels are built. In Elliptic, a transaction is labeled licit or illicit according to the category of the entity that initiated it (Weber et al., 2019). In Elliptic2, a subgraph is labeled suspicious when it connects a cluster that the data provider labeled illicit to a licit service, and licit when both ends are licit (Bellei et al., 2024). In both cases the label is a function of who sends the funds. Under a random split, a model can then score well by recognizing senders rather than laundering behavior, provided the same senders appear in training and test. Whether they do is not a property of the label rule but of the data: it depends on what a "sender" is and how often it recurs.

We therefore audit both benchmarks with one protocol. On Elliptic2, the shortcut is severe: a lookup of sender history with no learning nearly matches the published state of the art, and the subgraph's own structure carries limited signal that transfers weakly to new senders. Similar entity-level dependence between training and test data was recently documented for the original Elliptic dataset (Šafář et al., 2026), and Buu & Kim (2026) reported that Elliptic2 performance drops under an entity-disjoint split; we explain why. On Elliptic++ transaction classification, the same protocol finds the shortcut negligible: sender history adds almost nothing to a model of the transaction's own features, and those features keep most of their performance when sending addresses are held out.

Our claim is deliberately narrow. We do not argue that models trained on either benchmark are useless: recognizing the type of entity that sends funds is a real and transferable skill. We argue that on Elliptic2 this is what the benchmark mostly measures, so published scores there say little about whether a model has learned the structure of laundering itself, which is the capability the benchmark was introduced to test (Bellei et al., 2024). This is a form of shortcut learning (Geirhos et al., 2020), arising from how the labels were constructed. The Elliptic++ result shows that a sender-based label rule does not by itself imply such a shortcut, which is why the check has to be run on each benchmark rather than assumed.

**Contributions.**

1. **Reproduction.** We reimplement RevClassify (Song et al., 2024) from its description, reproduce its reported performance within 0.01 PR-AUC, and show that its performance rests on the sender set.
2. **Memorization baseline.** We show that a no-learning sender lookup nearly matches published state-of-the-art performance on Elliptic2's standard split.
3. **Entity-aware evaluation.** Building on the entity-disjoint evaluation of Buu & Kim (2026), we design a split for Elliptic2 that holds out sending entities while keeping every subgraph and balanced groups, with a diagnostic showing that it blocks memorization of suspicious sources.
4. **Novelty cue.** We identify a cue that appears only under entity-disjoint evaluation on Elliptic2, namely that an unseen sender predicts the suspicious class, and introduce a baseline that measures it.
5. **Decomposition.** On Elliptic2, we separate the signal carried by sender features, hub degree, receivers and the subgraph itself, with paired bootstrap intervals, and quantify how much of each transfers to unseen senders.
6. **Contrast on a second benchmark.** We apply the same protocol to Elliptic++ under random, temporal and sender-disjoint splits and show that the shortcut is negligible there, identifying the recurrence of sender identity, not the label rule alone, as the condition under which it arises.
7. **Protocol, recommendations and code.** We state the audit as a reusable protocol, give concrete evaluation recommendations, and will release code that rebuilds every result from the official downloads.

## 2 Related Work

**Graph learning for cryptocurrency AML.** Weber et al. (2019) released the first Elliptic dataset, a transaction graph in which nodes are labeled licit or illicit, and showed that graph convolutional networks and random forests could classify them; they also reported that all methods performed poorly after a dark-market shutdown late in the data's time span. Elmougy & Liu (2023) extended it as Elliptic++, adding 822,942 wallet addresses with 56 features each and the address-transaction graph, which enables both transaction and address classification. Elliptic2 (Bellei et al., 2024) changed both the granularity and the task: nodes are clusters of addresses believed to belong to one entity, and the prediction target is a subgraph. The dataset paper evaluated Sub2Vec, a GNN applied to each subgraph alone (GNN-Seg), and GLASS (Wang & Zhang, 2022), which uses the surrounding background graph. Song et al. (2024) proposed RevTrack, whose classifier RevClassify summarizes a subgraph by its senders and receivers with a DeepSets architecture (Zaheer et al., 2017) and reported a PR-AUC of about 0.97. Akter et al. (2026) benchmarked tree ensembles on Elliptic2 using features of each subgraph's own nodes and edges, reporting a PR-AUC of about 0.49 for a random forest. Buu & Kim (2026) proposed a reinforcement-learning traversal detector, reported an average precision of 0.966 under cross-validation, and noted that 97% of their evaluation subgraphs share at least one participant with the training data; under an entity-disjoint split, their detector reached 0.673.

**Leakage, shortcuts and evaluation.** Leakage, in which information unavailable at prediction time or correlated with the target enters training, is a recurring cause of overstated results (Kaufman et al., 2012; Kapoor & Narayanan, 2023). More broadly, models often succeed through shortcuts, features that predict the label on a benchmark without capturing the intended capability (Geirhos et al., 2020). In the digital-forensics context, Šafář et al. (2026) reverse-engineered most features of the original Elliptic dataset and documented temporal leakage, preprocessing leakage, and nonindependence between training and test samples arising from addresses that recur across splits. They recommended address-disjoint splits but did not analyze Elliptic2, whose task, unit of analysis and labeling rule differ. Buu & Kim (2026) observed both the participant overlap and the drop under entity-disjoint evaluation on Elliptic2, but did not explain the drop or identify where the remaining signal comes from. To our knowledge, no prior work has traced Elliptic2's evaluation behavior to its label construction, measured it with memorization and novelty baselines, decomposed it by feature block, or compared the strength of the sender shortcut across benchmarks that share a sender-based label rule. These are the contributions of this paper; the entity-disjoint result of Buu & Kim (2026) is consistent with ours and corroborates it independently.

## 3 Two Benchmarks, One Label Rule

### 3.1 Elliptic2

We use Elliptic2 as released by Elliptic on Kaggle (license CC BY-NC-ND 4.0). Table 1 lists the properties we verified directly from the files. Bellei et al. (2024) describe the construction as follows. Clusters are labeled licit (for example, exchanges and wallet providers), illicit (for example, dark markets and scams), or unknown, using heuristics and analyst judgment; these node labels are not released. Paths are traced from labeled clusters for up to six hops. A path from an illicit cluster to a licit one is suspicious; a path between licit clusters is licit. Subgraphs are formed from logically consistent paths, and only the unknown intermediate clusters are kept inside each subgraph. As Song et al. (2024) note, a subgraph is therefore suspicious when its senders are illicit and its receivers licit, and licit when both are licit. The sending entity sits just outside the subgraph, and its hidden label determines the subgraph's label.

This design has a direct consequence. Receivers are licit in both classes, so they carry little label information; senders carry nearly all of it. Any model that can recognize the sender, by identity or by features that reveal its type, can recover the labeling rule without learning anything about how funds move inside the subgraph.

### 3.2 Elliptic and Elliptic++

In the Elliptic transaction dataset, a transaction is deemed licit or illicit according to whether the entity initiating it belongs to a licit or an illicit category (Weber et al., 2019). The initiating entity is the one controlling the transaction's inputs, so the label is again defined on the sender. Elliptic++ (Elmougy & Liu, 2023) keeps the 203,769 transactions and their labels and adds, for each transaction, its input (sending) and output (receiving) addresses, together with per-address features for each time step. Table 1 lists the properties of the release we used.

The two benchmarks thus share a sender-based label rule but differ in what a sender is. In Elliptic2, a sender is an entity cluster, the result of grouping many addresses that are believed to belong to one actor; a few large services feed thousands of labeled subgraphs. In Elliptic++, a sender is an individual address, and Bitcoin addresses are often used only once or a few times. A memorization shortcut needs the same sender to appear in training and in test. Our hypothesis is that this condition holds strongly in Elliptic2 and weakly in Elliptic++, so the same label rule should produce shortcuts of very different strength. Sections 5 and 6 test it.

**Table 1:** Properties of the two benchmarks used in this study. Elliptic2 properties were verified from the released files; Elliptic++ label counts and sender coverage come from our processing of the release, and graph sizes and feature counts from its documentation. Elliptic++ counts refer to transactions with a licit or illicit label.

| **Property** | **Elliptic2** | **Elliptic++ (transactions)** |
|---|---|---|
| Unit classified | Subgraph of entity clusters | Transaction |
| Labeled units | 121,810 | 46,564 |
| Positive class | 2,763 suspicious (2.27%) | 4,545 illicit (9.76%) |
| Label defined on | Sending entity cluster | Initiating (sending) entity |
| Sender as given in the data | Entity cluster outside the subgraph | Input address of the transaction |
| Background graph | 49,299,864 clusters; 196,215,606 edges | 203,769 transactions; 234,355 edges |
| Units with an identified sender | 100% | 98.9% |
| Unit features | 43 node and 95 edge features (binned) | 183 transaction and 56 address features (as documented) |
| Time information | Not used | 49 time steps |

## 4 Evaluation Design

The protocol has four parts, applied to each benchmark: (i) a memorization baseline that scores a unit by the training history of its senders, with no access to the unit's own content; (ii) a split that holds out senders, verified by a leakage diagnostic; (iii) a novelty baseline, where the split makes it meaningful; and (iv) a decomposition of a strong learned model into feature blocks with and without sender information. Sections 4.1 to 4.5 describe the Elliptic2 implementation and Section 4.6 the Elliptic++ implementation.

### 4.1 Elliptic2: neighborhood extraction

The full edge file needs about 143 GB of memory with features loaded, so we stream both background files once and keep every edge that touches a node of a labeled subgraph, together with the features of every node those edges reach. This one-hop neighborhood has 737,870 nodes and 1,922,546 edges (about 1.6 GB with all features). All 367,137 internal subgraph edges are recovered. Transaction identifiers are not unique across edges, so edge features are joined on the source, destination and transaction identifier together.

## 4.2 Elliptic2: senders and receivers

Following the definitions of Song et al. (2024), a source is a subgraph node with no incoming edge from inside the subgraph and a sink is one with no outgoing edge to inside it. A sender is a node outside the subgraph with an edge into one of its sources; a receiver is a node outside the subgraph reached from one of its sinks. We assume edges point in the direction of fund flow. RevTrack's preprocessing code has not been released, so our extraction is an independent implementation. For 4,880 subgraphs without a source and 3,542 without a sink (because of cycles), we treat all subgraph nodes as sources or sinks respectively; Section 5.6 shows that removing these subgraphs instead leaves the conclusions unchanged.

The extraction is consistent with the direction assumption: a sender lookup is highly predictive while a receiver lookup is not (Section 5.1), matching the label rule, under which senders differ by class and receivers do not. Every subgraph has at least one sender; 95.5% have at least one receiver. The ten busiest senders together touch 46% of all subgraphs, and nearly all of their subgraphs are licit.

## 4.3 Elliptic2: splits

**Random.** The standard protocol: a random 80:10:10 split of subgraphs into training, validation and test sets (Bellei et al., 2024; Song et al., 2024).

**Entity-aware.** Subgraphs that share a sender are kept on the same side of the split. A strict version that groups by every sender is not usable: high-volume licit services connect almost all subgraphs into one group. We therefore exclude from grouping only hub senders that feed at least a cut-off number of subgraphs, and always keep in grouping any sender whose suspicious rate is at least a protection threshold (50% in the main analysis). The cut-off is the largest value for which no group exceeds 10% of subgraphs; with a 50% threshold it is 100, excluding 179 hubs. Groups are assigned to training, validation and test so that each split approaches 80:10:10 in both overall size and number of suspicious subgraphs.

**Verification.** For each test set we report the share of suspicious test subgraphs that have a sender which fed a suspicious subgraph in training. It is 0.907 under the random split (5 seeds) and 0.065 under the entity-aware split (20 seeds), confirming that the entity-aware split blocks memorization of suspicious sources.

## 4.4 Elliptic2: baselines and models

- **Lookup** (no learning): the smoothed suspicious rate of a subgraph's senders in the training set, taking the maximum over its senders; unseen senders receive the prior.
- **Novelty** (no learning): the share of a subgraph's senders that never appear in training.
- **Gradient boosting** (histogram-based, class-balanced; number of rounds chosen on validation PR-AUC) on five feature sets built by aggregating node and edge features with means and maxima: all features; sender features only; degree features only (how many subgraphs a sender or receiver touches, which is label-free); intrinsic features (internal nodes, internal edges, structure and receivers, with no sender information); and internal features only. No entity identifier is ever used as a feature. In total, 456 features.
- **RevClassify-DS** (reimplemented): each subgraph is represented only by the binned node features of its senders and of its receivers; each set passes through its own two-layer network, is mean-pooled, and is combined with the log set sizes by a classifier (DeepSets; Zaheer et al., 2017). It differs from the original in using the released features only (not node embeddings), our own sender and receiver extraction, and mean rather than sum pooling. A receivers-only variant removes all sender information. Before use, the implementation passed a positive control: on a small test graph with a planted sender signal, it learned the signal, and the receivers-only variant did not.

### 4.5 Metrics and uncertainty

We report PR-AUC (average precision) as the main metric, with ROC-AUC where useful. Because splits change the share of positive units in the test set, and because the two benchmarks differ greatly in prevalence (2.3% against 9.8%), we also report PR-AUC as a multiple of the test prevalence and the normalized lift (PR-AUC − $\pi$)/(1 − $\pi$), where $\pi$ is the test prevalence; the normalized lift is 0 for a random ranking and 1 for a perfect one, and is comparable across prevalences. On Elliptic2, entity-aware test sets contain more suspicious subgraphs on average (3.2% against 2.3%), which raises expected PR-AUC, so drops in PR-AUC under that split are conservative. Elliptic2 results are averaged over random seeds, with 95% stratified bootstrap intervals (300 resamples per seed) for each model and paired intervals for differences between models on the same test set. Elliptic++ results are averaged over five seeds and reported with the standard deviation across seeds; differences between models are reported per seed on the same test set.

### 4.6 Elliptic++: protocol

**Task.** Binary classification of labeled Elliptic++ transactions as illicit or licit; transactions of unknown class are excluded from training and evaluation. The sender of a transaction is the set of its input addresses, taken from the address-transaction edge list, and its receivers are its output addresses. At least one input address is identified for 98.9% of labeled transactions. We do not use the released address labels, so that no address-level label information can enter any model.

**Splits.** (i) *Random*: a stratified 70:30 split, varied over five seeds. (ii) *Temporal*: training on time steps 1–34 and testing on 35–49, the split used by Weber et al. (2019); it is fixed, and its seeds vary only the model and cross-fitting folds. (iii) *Sender-disjoint*: transactions that share any input address are joined into sender clusters by union-find, and whole clusters are assigned to training or test with a group shuffle that places 30% of clusters in test (28.6% of transactions on average), varied over five seeds. This gives 25,433 clusters, the largest holding 9.4% of labeled transactions. Unlike the Elliptic2 entity-aware split, no hub exclusion is needed, because no cluster holds more than 9.4% of transactions.

**Verification.** For each split we report the share of test transactions with at least one input address that also appears as an input address of a labeled training transaction. It is 0.468 under the random split, 0.266 under the temporal split and 0 by construction under the sender-disjoint split.

**Models.** All models use histogram-based gradient boosting without class weighting (300 rounds at most, learning rate 0.05, early stopping on an internal 10% validation fraction), with the classification threshold for F1 chosen on training predictions. Feature sets:

- **Sender history** (the counterpart of the Elliptic2 lookup): for each input address, the smoothed share of illicit transactions among the labeled training transactions it sent; the mean and the maximum over input addresses, plus an indicator that any input address was seen in training. Values for training rows are computed out of fold (five folds), so a transaction's own label never enters its value. Because a gradient-boosting model is fitted on these three features, this baseline is, if anything, stronger than the no-learning Elliptic2 lookup.
- **Full**: all released transaction features except the identifier and the time step (local, one-hop aggregate and augmented features).
- **Local**: the 93 local transaction features only, without neighborhood aggregates or address information; the closest counterpart of the Elliptic2 internal features.

- **Sender features** and **receiver features**: the mean and maximum of the address features of the input or output addresses at the transaction's time step (falling back to the address's mean over time steps), and the number of such addresses.
- **Full + sender history**: the union, which measures what sender history adds to the transaction's own features.

A positive control on a synthetic dataset with the same schema and a planted sender signal confirmed that the sender-history baseline recovers such a signal under random and temporal splits and falls to chance under the sender-disjoint split.

## 5 Results on Elliptic2

Random-split results are means over 5 seeds; entity-aware gradient-boosting results are means over 20 seeds with a 50% protection threshold. RevClassify-DS results use 5 seeds per split.

**Table 2:** Elliptic2: mean test PR-AUC, random split over 5 seeds and entity-aware split over 20 seeds. Prevalence is the mean share of suspicious subgraphs in the test sets. RevClassify-DS rows use 5 seeds per split. The reported RevClassify value is from Song et al. (2024).

| Model | Random | Entity-aware | Information used |
|---|---|---|---|
| Prevalence (chance) | 0.023 | 0.032 | — |
| Lookup (no learning) | 0.954 | 0.413 | Sender identity |
| Novelty (no learning) | 0.084 | 0.462 | Sender unseen |
| Gradient boosting, all | 0.975 | 0.746 | All blocks |
| Gradient boosting, sender | 0.969 | 0.749 | Sender features |
| Gradient boosting, degree | 0.867 | 0.560 | Hub degree |
| Gradient boosting, intrinsic | 0.619 | 0.186 | No sender information |
| Gradient boosting, internal | 0.559 | 0.136 | Subgraph only |
| RevClassify (reported) | 0.974 | — | Senders and receivers |
| RevClassify-DS (ours) | 0.966 | 0.623 | Senders and receivers |
| DeepSets, receivers only (ours) | 0.126 | 0.074 | Receivers only |

### 5.1 Under the standard split, memorization nearly matches the state of the art

Under the random split (Table 2), the lookup reaches 0.954 and the best learned model 0.975; the paired difference is 0.003 to 0.037. Ninety-one percent of suspicious test subgraphs have a sender that fed a suspicious subgraph in training. A simple gradient-boosting model also matches the reported RevClassify result without any graph neural network, which suggests that the published gains are driven by the input information, the sender, rather than the architecture. Our internal-only model (0.559) is close to the random-forest result of Akter et al. (2026), about 0.49, who aggregated features of each subgraph's own nodes and edges. This is consistent with our account: the large gap to the 0.97 range opens only when sender information is used.

### 5.2 Holding out senders

Under the entity-aware split, all-feature performance drops from 0.975 to 0.746 and the lookup from 0.954 to 0.413. Sender features remain the strongest block (0.749), and models using them exceed the lookup by a clear margin (paired difference, all features minus lookup: 0.29 to 0.38), indicating that sender features generalize to unseen senders. The size of the drop agrees with the entity-disjoint result of Buu & Kim (2026), 0.673, obtained with a different detector and a different split construction. We interpret this as learning what kind of entity the sender is, which is exactly the quantity that the label rule encodes, rather than learning the shape of laundering. Results vary widely across seeds: the all-feature PR-AUC ranges from about 3 to 78 times the test prevalence, depending on which large illicit sources fall in the test set. Performance on unseen criminals therefore depends strongly on which criminals they are, and single-split results on this benchmark should not be trusted.

### 5.3 The novelty cue

A sender that never appears in training is itself predictive of the suspicious class under the entity-aware split (novelty PR-AUC 0.462, against 0.084 under the random split). Licit subgraphs are typically fed by large services that also appear in training, while suspicious subgraphs are fed by smaller illicit clusters, which an entity-aware split makes unseen by construction. In a stricter split that grouped by every sender connected to a suspicious subgraph, no suspicious test subgraph had any sender seen in training, yet the lookup still reached 0.866, because unseen senders received the prior and seen licit hubs received lower scores. Any entity-disjoint evaluation on this dataset should therefore report the novelty baseline as a floor.

### 5.4 How much of the laundering pattern transfers?

Without any sender information, performance falls from 0.619 under the random split to 0.186 under the entity-aware split, about six times the test prevalence of 0.032; the strict split gives a similar ratio (0.117 against a test prevalence of 0.017). Receivers add a small but consistent amount (intrinsic minus internal: 0.03 to 0.09 under the random split and 0.03 to 0.08 under the entity-aware split). The structure and transactions of a laundering subgraph therefore carry real but limited information about new criminal sources, and the random split overstates it about threefold in raw PR-AUC and about fivefold relative to prevalence.

### 5.5 The published architecture behaves the same way

Our RevClassify-DS reimplementation reaches 0.966 under the random split, reproducing the reported 0.974 without node embeddings. Under the entity-aware split it falls to 0.623; on the same five splits, gradient boosting on all features reaches 0.684. The drop is therefore not specific to our baseline. Removing senders from the same architecture leaves 0.126 (random) and 0.074 (entity-aware), only about three to five times prevalence: receivers alone carry little information, as the label rule predicts, since they are licit in both classes. The near-perfect published result rests on the sender set.

### 5.6 Sensitivity

Changing the protection threshold from 50% to 30% or 70% changes the protected set only among senders below the hub cut-off, which are grouped regardless. The 30% and 70% runs (10 seeds each) produced identical splits and identical results, and both agree with the 50% run within seed variation (all features: 0.732 against 0.746; intrinsic: 0.191 against 0.186). The conclusions therefore do not depend on this threshold.

We also varied the hub cut-off (50 and 200, against 100 in the main analysis) and removed the 5,596 subgraphs without a clear source or sink instead of applying the fallback rule (10 entity-aware seeds each; 5 random seeds for the reduced dataset). Table 3 summarizes the results. In every configuration the split blocks memorization (5% to 6% of suspicious test subgraphs have an informative seen sender, against 92% under the random split), sender-based models stay far above the subgraph's intrinsic signal, and the intrinsic signal stays between 0.180 and 0.191, within 0.01 of the main analysis. On the reduced dataset, the random-split results reproduce the full-data values almost exactly (all features 0.978 against 0.975; lookup 0.958 against 0.954), so the fallback rule for cyclic subgraphs does not drive any result. The level of sender-based performance varies more across configurations (0.66 to 0.80), as expected given the large seed-to-seed variation reported in Section 5.2.

**Table 3:** Elliptic2: sensitivity of mean test PR-AUC to the hub cut-off and to the handling of cyclic subgraphs. Rows other than the main analysis use 10 entity-aware seeds; the last row uses 5 random seeds.

| Configuration | Prevalence | All features | Sender | Intrinsic | Lookup / Novelty |
|---|---|---|---|---|---|
| Main: cut-off 100 (20 seeds) | 0.032 | 0.746 | 0.749 | 0.186 | 0.413 / 0.462 |
| Cut-off 50 | 0.033 | 0.779 | 0.778 | 0.185 | 0.455 / 0.526 |
| Cut-off 200 | 0.027 | 0.802 | 0.799 | 0.191 | 0.362 / 0.398 |
| Cyclic subgraphs removed | 0.055 | 0.656 | 0.655 | 0.180 | 0.329 / 0.352 |
| Cyclic removed, random split | 0.023 | 0.978 | 0.972 | 0.626 | 0.958 / 0.077 |

# 6 Results on Elliptic++

Table 4 reports the Elliptic++ results under the three splits. All numbers are means over five seeds; Table 5 gives the per-seed values for the sender-disjoint split, where the variation is large.

**Table 4:** Elliptic++ transaction classification: mean test PR-AUC (± standard deviation over 5 seeds) and, in parentheses, PR-AUC as a multiple of test prevalence. Prevalence is the mean share of illicit transactions in the test sets.

| Feature set | Random | Temporal | Sender-disjoint |
|---|---|---|---|
| Prevalence (chance) | 0.098 | 0.065 | 0.105 |
| Test units with a sender seen in training | 0.468 | 0.266 | 0 |
| Sender history (lookup features) | 0.340 ± 0.009 (3.5×) | 0.091 ± 0.000 (1.4×) | 0.107 ± 0.024 (1.0×) |
| Full transaction features | 0.986 ± 0.002 (10.1×) | 0.808 ± 0.003 (12.4×) | 0.892 ± 0.118 (8.5×) |
| Local transaction features | 0.980 ± 0.002 (10.0×) | 0.786 ± 0.002 (12.1×) | 0.883 ± 0.118 (8.4×) |
| Sender address features | 0.956 ± 0.003 (9.8×) | 0.379 ± 0.016 (5.8×) | 0.854 ± 0.110 (8.1×) |
| Receiver address features | 0.965 ± 0.004 (9.9×) | 0.635 ± 0.024 (9.8×) | 0.866 ± 0.113 (8.2×) |
| Full + sender history | 0.989 ± 0.002 (10.1×) | 0.812 ± 0.003 (12.5×) | 0.897 ± 0.109 (8.5×) |

## 6.1 Sender history carries little information

Under the random split, sender history reaches a PR-AUC of 0.340, or 3.5 times the test prevalence, against 0.986 for the full transaction features (10.1 times). In normalized lift it recovers 0.268, against 0.985 for the full model, so it accounts for about 27% of the full model's gain over chance; on Elliptic2 the corresponding lookup accounts for 98% (Table 6). Adding sender history to the full feature set raises PR-AUC by 0.0019 to 0.0035 in every seed (paired on the same test set) and changes it by −0.0003 to +0.008 under the temporal split. The transaction's own features already contain nearly all the information that sender history provides.

The reason is visible in the verification diagnostic. Under the random split, only 46.8% of test transactions have any input address that also sent a labeled training transaction, and 26.6% under the temporal split. On Elliptic2, 91% of suspicious test subgraphs had a sender that had fed a suspicious subgraph in training. A memorization shortcut cannot dominate when most test senders are new.

### 6.2 Holding out sending addresses

Under the sender-disjoint split, sender history falls to chance (0.107 against a prevalence of 0.105), as it must when no test sender is seen in training, and the indicator of a seen sender is constant, so the novelty cue of Section 5.3 cannot arise. The transaction features, however, keep most of their performance: the full model falls from 0.986 to 0.892 and the local features from 0.980 to 0.883, a normalized lift of 0.869, against 0.159 for Elliptic2's intrinsic features under its entity-aware split. The transaction's own content therefore transfers to unseen senders on Elliptic++, whereas the content of an Elliptic2 subgraph transfers weakly.

As on Elliptic2, results vary strongly with the split (Table 5). In three seeds the full model reaches 0.974 to 0.981; in two it falls to 0.775 and 0.750, and every feature set falls with it, while sender history stays at chance throughout. The drop is therefore not caused by sender memorization; it most likely reflects which sender clusters fall in the test set. We did not isolate the cause further, but single-split results should not be trusted on this benchmark either.

**Table 5:** Elliptic++ sender-disjoint split: test PR-AUC by seed.

| Feature set | Seed 0 | Seed 1 | Seed 2 | Seed 3 | Seed 4 |
|---|---|---|---|---|---|
| Sender history | 0.110 | 0.085 | 0.134 | 0.126 | 0.080 |
| Full transaction features | 0.974 | 0.981 | 0.775 | 0.750 | 0.980 |
| Local transaction features | 0.964 | 0.970 | 0.765 | 0.742 | 0.974 |
| Full + sender history | 0.975 | 0.977 | 0.789 | 0.769 | 0.978 |

### 6.3 The temporal split

Under the temporal split, the full model falls to 0.808 and sender features to 0.379. This decline is expected and is not a sender effect: Weber et al. (2019) reported that all methods perform poorly after a dark-market shutdown in the later time steps of the test period, which changes the mix of illicit activity. Sender history is near chance (1.4 times prevalence), because only a quarter of test transactions have a sender seen in the training period.

### 6.4 A caution on address features

Sender address features score highly under the random and sender-disjoint splits (0.956 and 0.854), but so do receiver address features (0.965 and 0.866). We do not read this as a sender effect. Elliptic++ address features summarize an address's transaction activity in a time step, and that activity can include the transaction being classified, so address features partly re-encode the transaction itself. They are also the less robust signal: under the temporal split they fall to 0.379 (sender) and 0.635 (receiver), against 0.808 for the transaction features. The sender-history baseline, which uses only labels of other training transactions, is the clean measure of the shortcut and is the one we compare across benchmarks.

## 7 Comparing the Two Benchmarks

Table 6 places the main quantities side by side. On the standard random split, a no-learning sender lookup on Elliptic2 attains 98% of the full model's gain over chance; on Elliptic++, a learned model on sender history attains 27%. When senders are held out, the signal that does not depend on the sender retains a normalized lift of 0.159 on Elliptic2 and 0.869 on Elliptic++. Both benchmarks define their labels on the sender. What differs is whether the sender recurs: Elliptic2 senders are entity clusters, and a few high-volume services feed thousands of subgraphs, whereas Elliptic++ senders are addresses, and most test transactions have no sending address seen in training (53% under the random split, 73% under the temporal split). The strength of a sender shortcut therefore depends on the recurrence of sender identity across evaluation units, which is a property of how the data are aggregated, not of the label rule alone.

**Table 6:** The sender shortcut on the two benchmarks. Lift share is (PR-AUC of the sender baseline − $\pi$)/(PR-AUC of the full model − $\pi$). Normalized lift is (PR-AUC − $\pi$)/(1 − $\pi$). Held-out split: entity-aware for Elliptic2 (20 seeds), sender-disjoint for Elliptic++ (5 seeds). Sender-free signal: intrinsic features for Elliptic2, local transaction features for Elliptic++.

| Quantity | Elliptic2 | Elliptic++ |
|---|---|---|
| Random split: sender baseline PR-AUC | 0.954 (41.5× prevalence) | 0.340 (3.5× prevalence) |
| Random split: full model PR-AUC | 0.975 (42.4×) | 0.986 (10.1×) |
| Random split: lift share of the sender baseline | 0.98 | 0.27 |
| Random split: leakage diagnostic | 0.907 (suspicious units with a seen suspicious sender) | 0.468 (all units with any seen sender) |
| Held-out split: full model, normalized lift | 0.738 | 0.879 |
| Held-out split: sender-free signal, normalized lift | 0.159 | 0.869 |
| Held-out split: novelty cue | Present (0.462 vs. 0.032) | Absent by construction |
| Verdict | Severe sender shortcut | Negligible sender shortcut |

## 8 Discussion and Recommendations

**What these results do and do not show.** They show that Elliptic2 performance, as currently reported, is driven mainly by information about the sending entity, and that the structural signal of laundering transfers weakly to new sources. They do not show that subgraph structure is useless for AML in general, nor that the published methods are incorrect: RevClassify does what it was designed to do, and we reproduce it. The issue lies in what the benchmark, as constructed, can measure. The Elliptic++ results show the other side: a benchmark with the same kind of label rule can still measure what a transaction looks like, because its senders recur much less often.

For practitioners, the distinction matters. A model that recognizes known criminal entities is useful, but it duplicates what attribution databases already provide and fails exactly where AML most needs help: new actors. A model that recognizes laundering behavior would generalize to them. Elliptic2's standard protocol cannot tell these apart, and the reported near-perfect scores reflect mostly the former.

**A diagnostic protocol.** Because the shortcut depends on the data rather than the label rule alone, it cannot be assumed present or absent from a benchmark's documentation. We recommend that evaluations on any AML benchmark whose labels are defined on an entity:

- report results under a sender- or entity-disjoint split in addition to the standard split, and verify with a leakage diagnostic that informative senders do not cross the split;
- include a sender lookup baseline, and a novelty baseline wherever the split leaves some senders seen, since a model's value is its margin over these, not over chance;
- report performance with and without sender information, since the label rule is defined on the sender;
- disclose how senders and receivers are extracted and how cycles are handled, since results depend on it;
- report PR-AUC relative to test prevalence, or as normalized lift, when splits or benchmarks differ in class balance;

- report results over many seeds or splits, since on both benchmarks held-out performance depends strongly on which senders fall in the test set.

## 9 Limitations

- Our Elliptic2 sender and receiver extraction is an independent implementation; RevTrack's preprocessing is unpublished, and edge direction is inferred, though consistent with the label rule.
- On Elliptic2 we use gradient boosting on aggregated features and a reimplementation of RevClassify; other architectures, such as GLASS with neighborhood sampling, were not rerun. On Elliptic++ we use gradient boosting only and did not rerun published graph neural networks.
- The two audits are matched in design but not identical in implementation. The Elliptic2 lookup uses no learning, while the Elliptic++ sender history feeds lookup features to a gradient-boosting model; the Elliptic2 leakage diagnostic counts suspicious test units with a seen suspicious sender, while the Elliptic++ diagnostic counts all test units with any seen sender; Elliptic2 differences carry paired bootstrap intervals, while Elliptic++ differences are reported per seed; and the two use different split ratios (80:10:10 against 70:30) and class weighting (balanced against none). The first choice favors finding a shortcut on Elliptic++ and the second overstates sender overlap there, so neither works against the conclusion that the shortcut is negligible on that benchmark.
- The Elliptic2 entity-aware split depends on a hub cut-off and a protection threshold; we report sensitivity to both. On both benchmarks, results vary across seeds because large sources fall on different sides of the split; we did not isolate the cause of the two weak Elliptic++ sender-disjoint seeds.
- Elliptic2 node labels are hidden, so we cannot verify its label rule directly; we rely on its published description. On Elliptic++, sending addresses are not clustered into entities; an entity-level clustering of addresses could make senders recur more often and strengthen the shortcut, which we did not test.
- We did not evaluate a temporal split on Elliptic2. Both benchmarks are Bitcoin data, and we make no claim about other blockchains or about bank transaction networks.

## 10 Conclusion

Elliptic2 subgraph classification largely reduces to classifying the sender. Under the standard split, a lookup of who sent the funds almost matches the best published result, which we reproduce. When senders are held out, performance falls, a novelty cue emerges, and what survives comes mainly from recognizing the type of sending entity; the structure of the laundering subgraph itself transfers only weakly to new criminal sources. The same audit on Elliptic++, whose labels are also defined on the sender, finds the shortcut negligible: senders are addresses that recur far less often, sender history adds almost nothing, and the transaction's own features transfer to unseen senders. A sender-based label rule is thus not sufficient for a sender shortcut; whether one arises depends on how often sender identity recurs across the units being classified. Entity-aware evaluation with explicit memorization and novelty baselines should be run on each benchmark before its results are read as evidence of detecting laundering behavior.

## Broader Impact Statement

This work concerns the evaluation of AML models, not their deployment. Its main risk is misreading: the finding that published Elliptic2 scores mostly reflect sender information could be taken to mean that graph-based AML detection does not work, which our results do not show, and the Elliptic++ results show the opposite case. A second consideration is that knowing which signals models rely on could, in principle, inform evasion. We judge this risk low: the dependence on sender identity follows directly from the published labeling rules, and the datasets' features are binned or anonymized. We use only the public releases under their terms and release no derived data.

## Data and Code Availability

Elliptic2 is available from Elliptic under CC BY-NC-ND 4.0; in keeping with the license, we release no derived data. Elliptic++ is available from its authors' public repository (Elmougy & Liu, 2023). Code that rebuilds every result in this paper from the official downloads of both datasets will be released in a public repository upon publication and is available to reviewers on request.

## Declaration of Generative AI Use

The author conducted this study, obtained and processed both datasets, ran all experiments on the author's own hardware using the official releases, and approved the study design, analyses and claims. Every reported number comes from these runs. In the course of the work, the author used Claude (Anthropic), an AI assistant, for support with the language editing and writing, and Grammarly for proofreading. The author takes full responsibility for the content of the article.